\documentclass{article} %
\usepackage{iclr2027_conference,times}

\usepackage{amsmath,amsfonts,bm}

\def\eqref#1{equation~\ref{#1}}

\def\1{\bm{1}}

\DeclareMathAlphabet{\mathsfit}{\encodingdefault}{\sfdefault}{m}{sl}
\SetMathAlphabet{\mathsfit}{bold}{\encodingdefault}{\sfdefault}{bx}{n}

\usepackage{hyperref}
\usepackage{url}

\usepackage{colortbl}
\usepackage{pifont}
\usepackage{xcolor}
\usepackage{graphicx}
\usepackage{booktabs}
\usepackage{cleveref}
\usepackage{enumitem}

\usepackage{algorithm}
\usepackage{listings}
\usepackage{pythonhighlight}   %
  
\usepackage{caption}

\usepackage{pifont}
\newcommand{\cmark}{\ding{51}}
\newcommand{\xmark}{\ding{55}}

\newcommand{\std}[1]{{\scriptsize$\pm$#1}}

\newcommand{\RN}[1]{\uppercase\expandafter{\romannumeral #1}}

\newcommand{\ours}{Poincar3}

\newcommand{\mat}[1]{\mathtt{#1}}

\crefname{figure}{Fig.}{Figs.}
\crefname{table}{Tab.}{Tabs.}
\crefname{section}{Sec.}{Secs.}
\crefname{subsection}{Sec.}{Secs.}
\crefname{equation}{Eq.}{Eqs.}
\crefname{algorithm}{Algo.}{Algos.}

\title{Emergent Multi-View Geometry Through Self-Distillation}

\vspace{-5pt}
\author{\textbf{David Nordstr\"om$^{1}$,\quad Thibaut Loiseau$^{2}$,\quad Vincent Lepetit$^{2}$}\\
  \textbf{Michael Felsberg$^{3}$,\quad Guillaume Bourmaud$^{4}$,\quad Fredrik Kahl$^{1}$} \\
  $^1$Chalmers University of Technology, Sweden\\
  $^2$LIGM, Ecole des Ponts, Univ. Gustave Eiffel, CNRS, France\\ 
  $^3$Link\"oping University, Sweden\\
  $^4$Univ. Bordeaux, CNRS, Bordeaux INP, IMS, UMR 5218, France
}

\definecolor{studentcolor}{HTML}{4285f4}
\definecolor{teachercolor}{HTML}{d6582b}

\iclrfinalcopy %
\begin{document}

\maketitle

\begin{abstract}
Over a century ago, Henri Poincar\'e argued that a motionless observer cannot acquire the notion of space. Yet, most visual representation learning methods operate on individual images, while those that leverage multiple views rely on RGB reconstruction, entangling geometry with appearance.
We propose \ours, a self-supervised method that learns representations from multiple views through self-distillation instead of RGB reconstruction. We combine masked patch and image-level distillation with a teacher that observes additional views, enabling training from scratch without explicit 3D supervision. \ours~outperforms both previous single and multi-view self-supervised approaches such as DINOv3, MuM, and Muskie on correspondence estimation, camera pose estimation, and 3D reconstruction. Using a lightweight Poincar\'e adapter, we also find that our learned features encode camera motion more accurately than existing self-supervised representations. Code and weights available publicly at: \url{https://github.com/davnords/poincar3}.%
\end{abstract}
    
\begin{figure}[b]
\centering
\includegraphics[width=0.99\linewidth]{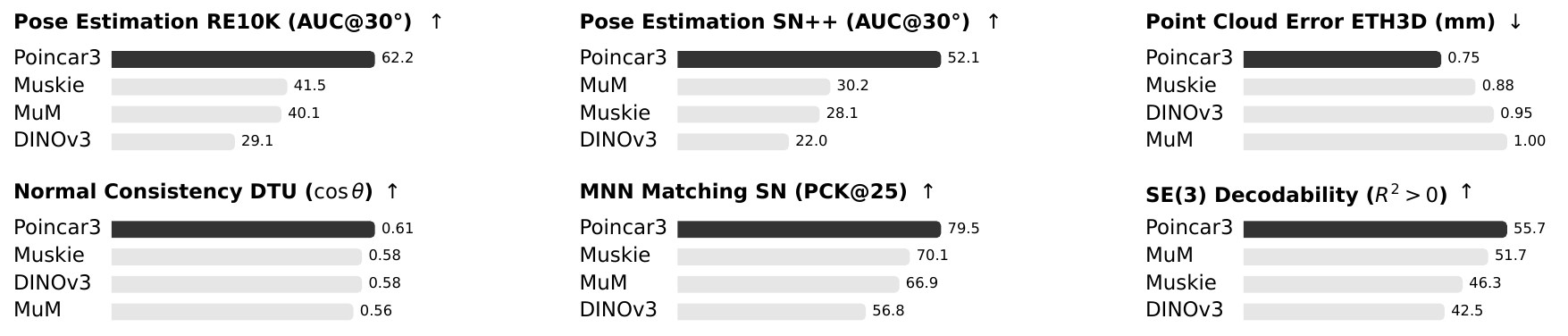}
\caption{\textbf{\ours~main results.} \ours~outperforms state-of-the-art SSL baselines across multi-view geometric tasks, including pose estimation, point cloud estimation, and image matching.}
\label{fig:teaser}
\end{figure}

\section{Introduction} \label{sec:intro}

Representation learning lies at the heart of modern computer vision. Self-supervised learning (SSL) has produced increasingly powerful single-image representations without requiring human annotations~\citep{caron2021dino,zhou2022ibot,he2022mae,oquab2023dinov2,simeoni2025dinov3,yang2026pursuit}. For 3D vision, however, as Poincar\'e argued~\citep{poincare1905value}, a single image is inherently limited. Recent SSL approaches therefore exploit multiple images of the same scene to encourage representations to capture information shared across views. CroCo~\citep{weinzaepfel2022croco,weinzaepfel2023croco}, MuM~\citep{nordstrom2026mum}, and Muskie~\citep{li2025muskiemultiviewmaskedimage} extend masked image modeling~\citep{he2022mae} to the multi-view setting, where information from one image can be used to predict content in another.%

However, RGB reconstruction imposes unnecessary constraints on learned representations. Reconstructing RGB values across views requires encoding appearance alongside geometry, whereas downstream 3D tasks benefit from representations robust to appearance variations. This motivates a natural question: \textit{can we learn to transfer information across views without RGB reconstruction?}

We address this question through self-distillation, which has proven effective for learning single-image representations~\citep{caron2021dino,oquab2023dinov2,simeoni2025dinov3}. Self-distillation, however, is notoriously difficult to train, and introducing a multi-view transformer only exacerbates this difficulty. Yet, we show that a multi-view teacher--student objective can be trained entirely from scratch, without any 3D supervision.

\begin{figure*}[t]
\centering
\includegraphics[width=0.99\linewidth]{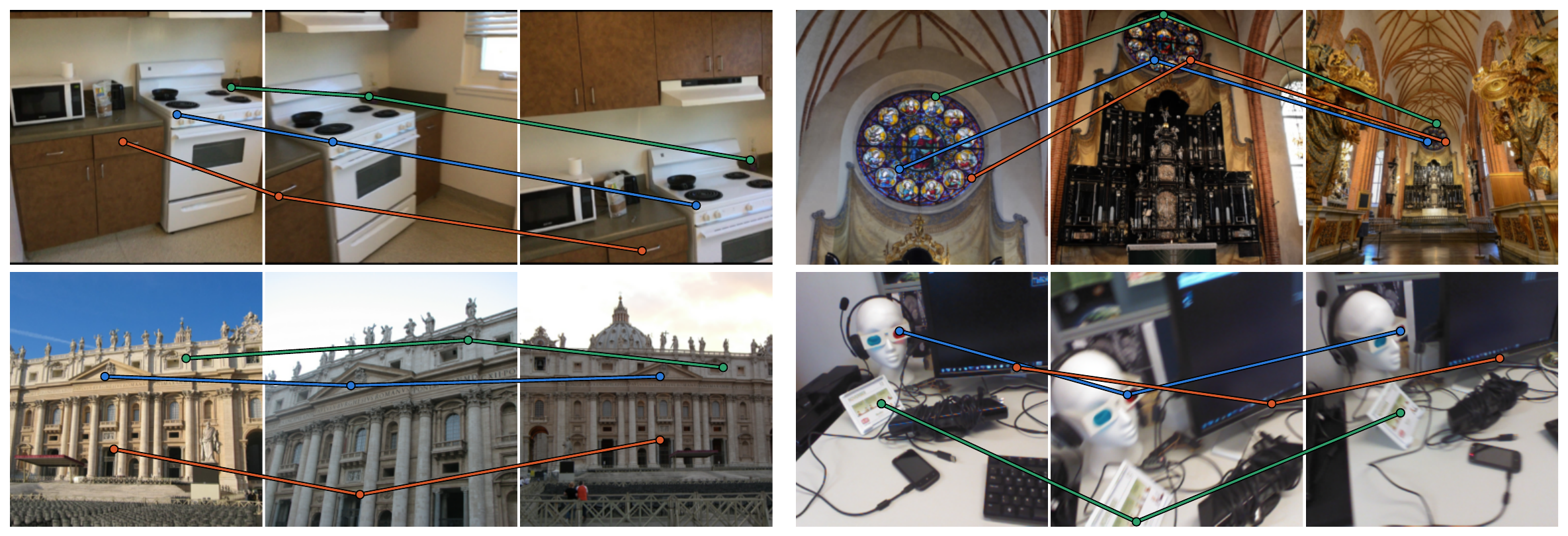}
\caption{\textbf{Emerging matching capabilities without supervision.} Given query keypoints, we visualize the tracks formed by selecting the patch with the highest attention activation. Despite receiving neither correspondence labels nor explicit attention supervision, the model learns to identify patch correspondences across images, suggesting the emergence of 3D-aware representations from SSL.}
\label{fig:attention-grid}
\end{figure*}

We find that stable and effective multi-view self-distillation relies on three key ingredients: introducing an image-level distillation objective, providing the teacher with access to additional views of the scene, and retaining full images rather than using local and global crops. 
Together, these ingredients give rise to \ours,\footnote{In reference to Poincar\'e's argument that understanding space requires motion. See also \citet{chen2026seese3emergence3dspace}.}a multi-view self-supervised model that learns strong representations directly from unlabeled image sequences mined from the internet. We illustrate its state-of-the-art empirical performance in \cref{fig:teaser}, and show that multi-view geometry emerges without labels, as evidenced by zero-shot correspondences in \cref{fig:attention-grid}.

Across a broad range of 3D tasks, such as relative pose estimation, point cloud estimation, and correspondence estimation, \ours~ substantially improves over state-of-the-art single- and multi-view SSL approaches, including DINOv3, MuM, and Muskie.

\noindent\textbf{The main contributions are as follows:}
\begin{enumerate}[label=(\roman*), topsep=0pt]
\item A new SSL objective based on multi-view self-distillation for learning visual representations from unlabeled image sequences, without RGB reconstruction or 3D supervision.
\item We show that the learned representations exhibit emergent multi-view geometric capabilities, including zero-shot multi-view correspondence estimation and lightweight $\mathrm{SE(3)}$  decodability via a Poincar\'e adapter, despite receiving no correspondence or 3D supervision.
\item An ablation study of the key components underlying our method.
\end{enumerate}

\section{Related work} \label{sec:related}

\paragraph{Self-Supervised Learning (SSL).} SSL aims to learn powerful representations from unlabeled data. In vision, early works used hand-crafted pretext tasks~\citep{doersch2015unsupervised,zhang2016colorful,gidaris2018unsupervisedrepresentationlearningpredicting}. Later work focused on clustering~\citep{Caron_2018_ECCV,asano2020self,caron2020unsupervised,caron2021dino} and contrastive learning~\citep{oord2018representationlearningcontrastivepredictive,misra2020selfsupervisedlearningpretextinvariantrepresentations,chen2020simple,he2020momentum}. A major advance was the exponential moving average (EMA) teacher--student framework~\citep{meanteachers:2017} which has since become the dominant SSL paradigm~\citep{grill2020bootstrap,he2020momentum,caron2021dino,zhou2022ibot,oquab2023dinov2,CAPI:2025,simeoni2025dinov3}. The current state-of-the-art, DINOv3~\citep{simeoni2025dinov3}, combines these advances with a masked image modeling objective based on iBOT~\citep{zhou2022ibot}. Concurrently with iBOT, \citet{he2022mae} introduced another masked image modeling objective (MAE) that instead reconstructs RGB pixels. Although simple and effective for pretraining, MAE-style models produce frozen representations less suited for efficient linear extraction than DINO-style methods~\citep{Zhou2026UnderstandingGR}.

\paragraph{SSL in 3D Vision.} To learn representations for 3D downstream tasks, CroCo~\citep{weinzaepfel2022croco,weinzaepfel2023croco} extended the MAE objective by conditioning the reconstruction on an unmasked reference view of the same scene, while Alligat0R~\citep{aligat0r:2025} and Gekko~\citep{loiseau2026revisiting} focused on covisibility segmentation. MuM~\citep{nordstrom2026mum} and Muskie~\citep{li2025muskiemultiviewmaskedimage} concurrently generalized CroCo to reconstruct an arbitrary number of views from the same scene using a multi-view transformer. \citet{zeroco:cvpr} showed that these models can perform zero-shot correspondence estimation. In this paper, we show that self-distillation substantially improves it. Other approaches include adapting existing foundation models to 3D~\citep{MultiViewFoundationModels2025}, using RGB-D guidance~\citep{almukhamedov2026dinocularselfsupervisedvisuospatialrepresentations}, novel view synthesis~\citep{jin2025lvsm,jiang2025rayzer,mitchel2026trueselfsupervisednovelview,zhao2026erayzer,lucas2026sparse}, and video models~\citep{kong2024hunyuanvideo,assran2025vjepa2,murlabadia2026vjepa2_1,wang2026genception}. An important downstream application is \textit{feed-forward reconstruction}, where a multi-view transformer directly predicts scene geometry from images. Recent models include~\citep{wang2024dust3r,leroy2024grounding,wang2025vggt,wang2025pi3,keetha2026mapanything,depthanything3,wang2026vggtomega}. VGGT-$\Omega$~\citep{wang2026vggtomega}, the strongest feed-forward reconstruction model to date, showed that post-training with an SSL objective enabled training on internet-scale videos, improving generalization. However, this approach first requires supervised pretraining with 3D annotations. In a similar spirit, SelfEvo~\citep{huang2026selfimproving4dperceptionselfdistillation} showed that VGGT can improve itself by using an EMA copy as a teacher, providing the teacher with additional views, and aligning teacher and student predictions. In contrast, we introduce a teacher--student objective that learns entirely from scratch without 3D annotations. It produces representations that outperform previous SSL baselines and are competitive with those obtained from supervised 3D training for zero-shot multi-view correspondence estimation.

\begin{figure*}[t]
\centering
\includegraphics[width=0.99\linewidth]{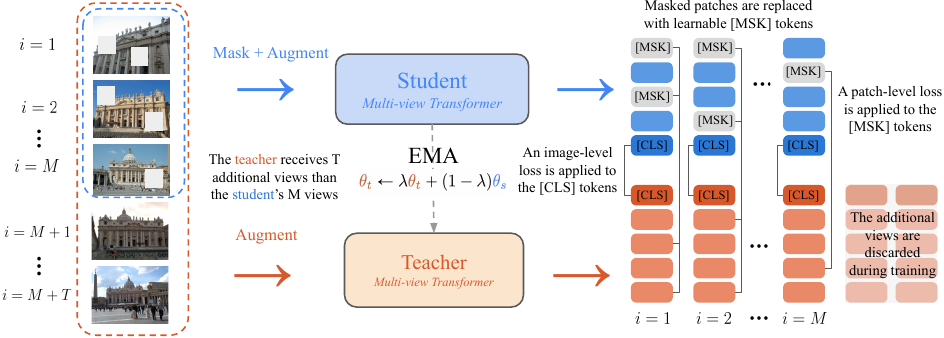}
\caption{\textbf{\ours} learns visual features without labels via multi-view self-distillation. The teacher, an EMA of the student, processes a full image sequence, while the student processes a masked and photometrically augmented subset. After the multi-view transformer, student and teacher embeddings are aligned using an image-level loss on CLS tokens and a patch loss on masked tokens.
}
\label{fig:method}
\end{figure*}

\section{Method}

\begin{figure}[t]
\centering
\begin{minipage}[t]{0.58\linewidth}
    \vspace{0pt}
    \captionsetup{skip=0pt}
    \captionof{algorithm}{Multi-view self-distillation pseudo-code.}
    \label{algo:multi-view-sd}
    \definecolor{codegreen}{rgb}{0.1,0.5,0.1}
    \lstset{
      basicstyle=\fontsize{7.2pt}{7.2pt}\ttfamily,
      commentstyle=\fontsize{7.2pt}{7.2pt}\color{codegreen}\itshape,
      keywordstyle=\fontsize{7.2pt}{7.2pt}\bfseries,
      upquote=true,
      showstringspaces=false,
      tabsize=4,
    }
    \begin{lstlisting}[language=python]
    # fs, ft: student and teacher networks
    # tps, tpt: student and teacher temperatures
    # l: EMA momentum rate
    # M, T: number of student and extra teacher views
    ft.params = fs.params
    for imgs in loader: # mini-batch of M+T frame seqs.
        sv = augment(imgs[:, :M])
        tv = augment(imgs)
        mask = sample_mask(sv) # patch mask [B, M, N]
    
        P_s, G_s, G_raw = fs(sv, mask=mask)
        with no_grad():
            P_t, G_t, _ = ft(tv, mask=None)
    
        # masked patch distillation (iBOT-style)
        P_t = sknopp(P_t[:, :M][mask].detach(), tpt)
        P_s = log_softmax(P_s[mask] / tps, dim=-1)
        L_patch = -(P_t * P_s).sum(-1).mean()
    
        # per-frame image objective (DINO-style)
        G_t = sknopp(G_t[:, :M].detach(), tpt)
        G_s = log_softmax(G_s / tps, dim=-1)
        L_global = -(G_t * G_s).sum(-1).mean()
    
        L_koleo = koleo(G_raw)
    
        loss = L_patch + 0.5*L_global + 0.1*L_koleo
        loss.backward()
        update(fs)  # AdamW
        ft.params = l*ft.params + (1-l)*fs.params
    \end{lstlisting}
\end{minipage}%
\hfill
\begin{minipage}[t]{0.40\linewidth}
    \vspace{0pt}
    \centering
    \includegraphics[width=\linewidth]{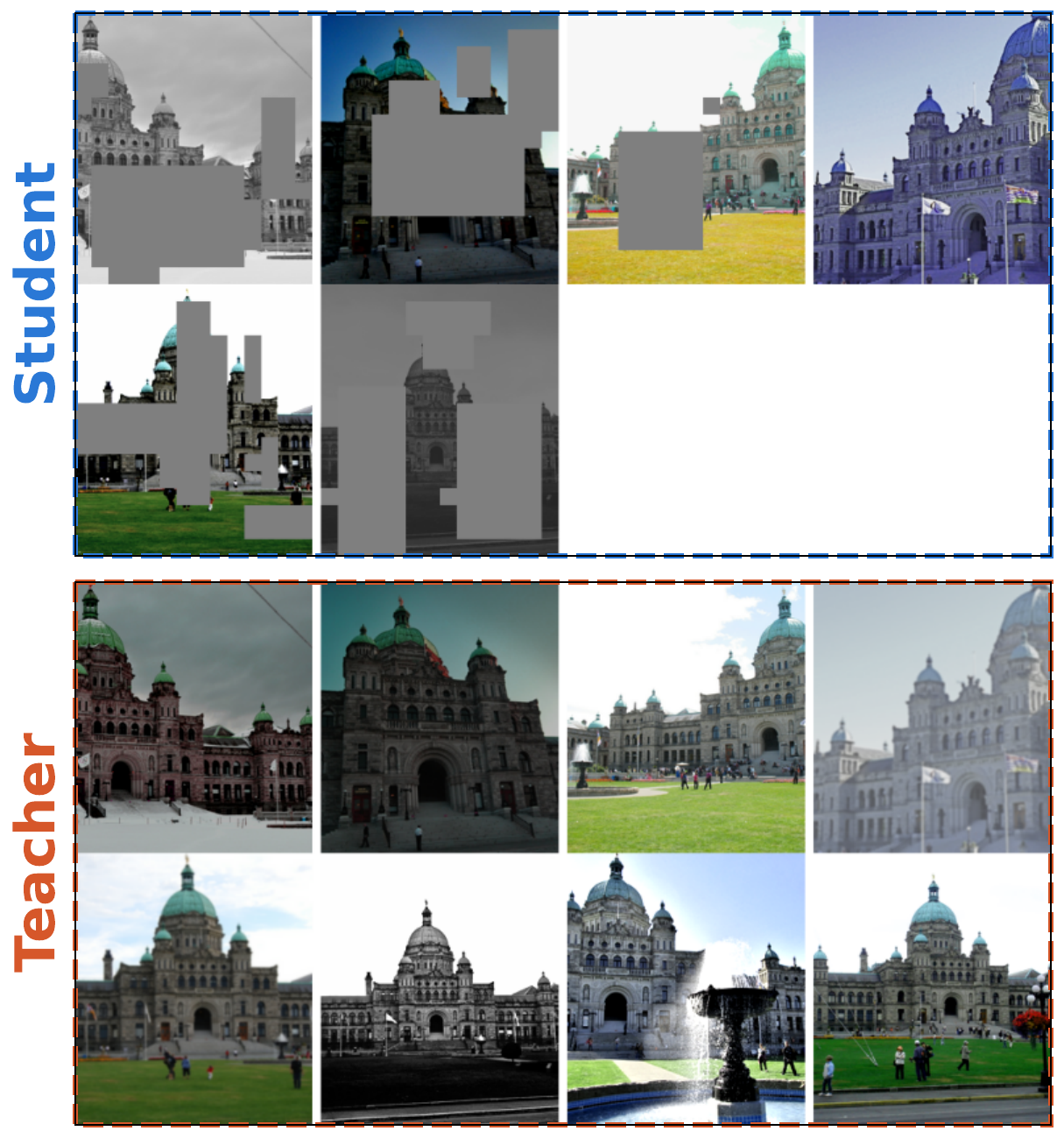}
    \caption{\textbf{Visualizing a training sequence.} The student and teacher get the same sequence (here $M=6$) while the teacher gets additional frames (here $T=2$), the student gets masked frames. %
    }
\label{fig:training-views}
\end{minipage}
\end{figure}

In this section, we begin by introducing the notation (\cref{subsec:notations}) and subsequently propose multi-view self-distillation (\cref{subsec:mvsd}). Thereafter, we outline the architecture (\cref{subsec:arch}) and training (\cref{subsec:training}) of \ours.  We illustrate \ours~in \cref{fig:method} and propose a pseudo-code implementation in \cref{algo:multi-view-sd}.

\subsection{Notations}\label{subsec:notations}

Let $\mathcal{I} = \{\mat{I}_1, \mat{I}_2, \dots, \mat{I}_M\}$ be a sequence of $M$ images of the same scene. Our goal is to learn a set of dense patch features $\mathcal{Z}=\{\mat{Z}_1,\ldots,\mat{Z}_M\}$, where $\mat{Z}_i\in\mathbb{R}^{N\times C}$ contains a $C$-dimensional feature vector for each of the $N$ patches in the $i$th image. These dense patch features should capture the underlying 3D structure of the scene, making them useful for downstream 3D vision tasks.

In practice, we parameterize our model by a backbone network $f_\theta$ that maps an image sequence to a set of dense patch features together with a global image representation $\mat{H}_i\in\mathbb{R}^{C}$ for each image:
    $(\mathcal{Z},\mathcal{H})=f_\theta(\mathcal{I})$,
where $\mathcal{H}=\{\mat{H}_1,\ldots,\mat{H}_M\}$ is the global (\texttt{[CLS]}) representation for each image.

\subsection{Multi-view Self-Distillation}\label{subsec:mvsd}

Our goal is to obtain a multi-view self-supervised learning method that does not rely on RGB reconstruction. To do so, we adopt the teacher--student self-distillation framework. The student network is parameterized by $\textcolor{studentcolor}{\theta_s}$ and the teacher by $\textcolor{teachercolor}{\theta_t}$. During training, the student learns to match the predictions of the teacher, while the teacher parameters are updated as an exponential moving average (EMA) of the student parameters $\textcolor{teachercolor}{\theta_t}
    \leftarrow
    \lambda\textcolor{teachercolor}{\theta_t}
    +
    (1-\lambda)\textcolor{studentcolor}{\theta_s}$, where $\lambda\in[0,1)$ denotes the EMA decay.

The current state of the art for visual self-distillation is DINOv3~\citep{simeoni2025dinov3}, built on the DINOv2~\citep{oquab2023dinov2} objective. The model operates on a single image by generating multiple augmented crops. The student processes masked global crops together with local crops, while the teacher observes only unmasked global crops. Both networks predict global and patch-level representations, and the student is trained to match the teacher using cross-entropy losses on the global representation and masked patches. Sinkhorn-Knopp~\citep{SinkhornKnopp1967} and KoLeo~\citep{sablayrolles2019spreading} regularization are applied to prevent representational collapse.

Although highly effective for single-image representation learning, we show in \cref{subsec:ablations} that naively extending this objective to multi-view image sequences leads to worse performance than RGB reconstruction. We therefore next consider the design of a multi-view self-distillation objective.

We operate on image sequences from the same scene where a multi-view transformer propagates information between frames. The backbone outputs are passed through a projection head (MLP + softmax) to transform the representation into a high-dimensional vector, as done in DINO, producing patch predictions
$\mat{P}_i=g_p(\mat{Z}_i)$,
and global predictions
$\mat{G}_i=g_g(\mat{H}_i)$.
For clarity, we denote the student outputs by $(\mathcal{P}^s,\mathcal{G}^s)$ and the teacher outputs by $(\mathcal{P}^t,\mathcal{G}^t)$.

\paragraph{Masked patch prediction.}

For every image $\mat{I}_i$, we randomly sample a binary patch mask \linebreak
$\mathbf{m}_i\in\left\{0,1\right\}^{N}$,
where $\mathbf{m}_i(u)=1$ indicates that patch $u$ of $\mat{I}_i$ is masked. We denote the corresponding set of masked patch indices 
by $\mathcal{M}_i$.
The student receives the masked image
$\tilde{\mat{I}}_i=(1-\mathbf{m}_i)\odot \mat{I}_i$,
where $\odot$ denotes masking at the patch level. The teacher always observes the full image.

Following iBOT~\citep{zhou2022ibot}, the patch prediction loss is computed only over masked patches,
\begin{equation}
\mathcal{L}_{\mathrm{patch}}
=
\frac{1}{\sum_{i=1}^M|\mathcal{M}_i|}
\sum_{i=1}^{M}
\sum_{u\in \mathcal{M}_i}
\mathrm{CE}
\left(
\mathcal{P}_{i,u}^{t},
\mathcal{P}_{i,u}^{s}
\right),
\end{equation}
where $|\mathcal{M}_i|$ denotes the number of masked patches, and 
    $\mathrm{CE}(\mathbf{p},\mathbf{q})\text{=}\,\text{-}\sum_k \mathbf{p}_k\log \mathbf{q}_k$
denotes the cross-entropy loss.

\begin{figure}[t]
\centering
\includegraphics[width=0.99\linewidth]{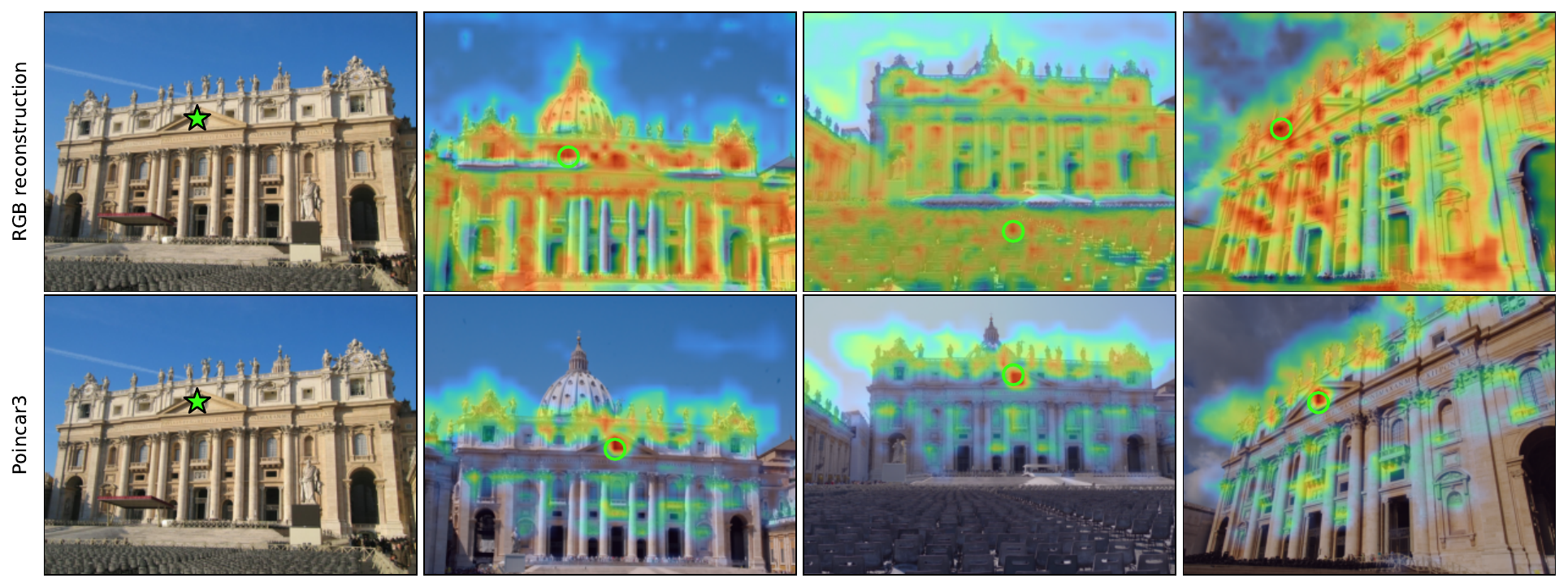}
\caption{\textbf{Qualitative feature comparison.} Given a query patch (green star), we visualize its feature correlations across frames and mark the maximum response (green circle). RGB reconstruction produces diffuse responses, whereas our method yields localized responses at corresponding patches.}
\label{fig:feature-corr}
\end{figure}

\paragraph{Image-level self-distillation.}

We additionally align the image-level representations using,
\begin{equation}
\mathcal{L}_{\mathrm{global}}
=
\frac{1}{M}
\sum_{i=1}^{M}
\mathrm{CE}
\left(
\mathcal{G}_i^t,
\mathcal{G}_i^s
\right),
\end{equation}
which we find essential for guiding self-distillation towards a 3D-aware representation.
\paragraph{Teacher with additional views.} We further propose that the teacher observe more of the scene than the student. If the student processes $M$ views, the teacher also receives $T$ extra frames from the same scene. These additional frames are excluded from all loss computations and only provide extra geometric context through the multi-view attention mechanism. During training, we sample $T$ uniformly between 0 and 12. We illustrate an example sequence in \cref{fig:training-views}.

\paragraph{Regularization and final objective.}

We apply Sinkhorn-Knopp and KoLeo regularization to prevent representational collapse. Our final training objective is,
\begin{equation}
\mathcal{L}
=
\mathcal{L}_{\mathrm{patch}}
+
\alpha
\mathcal{L}_{\mathrm{global}}
+
\beta
\mathcal{L}_{\mathrm{KoLeo}},
\end{equation}
where $\alpha=0.5$ and $\beta=0.1$ in all experiments.

\subsection{Network Architecture}\label{subsec:arch}

We parameterize $f_\theta$ as a multi-view transformer inspired by VGGT-$\Omega$~\citep{wang2026vggtomega}. The network consists of a ViT-L~\citep{dosovitskiy2021vit} image encoder followed by a multi-view transformer decoder. The decoder propagates information across views using alternating frame-wise and global attention. Rotary positional embeddings (RoPE)~\citep{rope:2023} are applied only within the frame-wise attention layers, while register attention~\citep{wang2026vggtomega} is employed in a subset of the decoder blocks to enable efficient scaling to long sequences. Unless otherwise stated, the decoder comprises $L=12$ alternating-attention layers with hidden dimension $C=1024$. In total, our model comprises around 650M trainable parameters.

\subsection{Training}\label{subsec:training}

\paragraph{Implementation Details.} We pre-train \ours~for 400k steps using the AdamW~\citep{adamw:2019} optimizer. Following VGGT and MuM, the input sequence length is sampled uniformly between 2 and 24 views while images are resized to $256\times256$. Given the sampled sequence length, we include as many training scenes as possible without exceeding a budget of 64 images per GPU. We use constant optimization hyper-parameters with learning rate $2\times10^{-4}$, weight decay $0.04$, and EMA decay $\lambda=0.999$. Training on 8 H200 GPUs takes three days. Further details on hyper-parameters are provided in \cref{subsec:supp:hparams} in the Appendix.

Downstream performance improves steadily throughout training (see \cref{fig:training-dynamics,fig:training-dynamics-detailed}), while our features become increasingly consistent across views compared to RGB reconstruction (see \cref{fig:feature-corr}). Interestingly, unlike DINOv3, we do not experience dense feature degradation during long training runs, which has been a topic of discussion lately~\citep{dai2025exploringstructuraldegradationdense,simeoni2025dinov3}.

\paragraph{Training Data.} Our method is fully self-supervised and requires only unlabeled image sequences. We train on large-scale collections of internet videos such as  SpatialVID~\citep{wang2025spatialvid} and RealEstate10K~\citep{realestate10k} and popular 3D annotated datasets such as MegaDepth~\citep{li2018megadepth} and ScanNet++~\citep{yeshwanth2023scannet++}. For the former, sequences are constructed via temporal random sampling, whereas for the latter, frames are randomly sampled from the same scene to form a sequence. In \cref{subsec:ablations}, we show the strength of being able to train on data lacking 3D annotations. The full data distribution is provided in \cref{tab:dataset_mix} in the Appendix.  

\section{Results}\label{sec:results}

In this section, we compare \ours's visual features to those of state-of-the-art SSL models on a wide range of 3D vision tasks. We begin by evaluating feed-forward reconstruction (\cref{subsec:reconstruction}), both through finetuning and using frozen features. Thereafter, we consider the ability to extract accurate correspondences from multiple views zero-shot (\cref{subsec:mv-match}) and decoding camera motion through a lightweight adapter (\cref{subsec:poincare}). Finally, we conduct extensive ablations that distinguish the contributions of each part of our design (\cref{subsec:ablations}). In the Appendix, we discuss our choice of baselines in \cref{sec:appendix:baselines}, additional experiments in \cref{sec:appendix:experiments}, and details on the evaluation in \cref{appendix:evaluation}. 

\subsection{Feed-Forward 3D reconstruction} \label{subsec:reconstruction}

\begin{figure}[t]
    \centering
    \begin{minipage}[t]{0.33\linewidth}
        \centering
        \includegraphics[width=\linewidth]{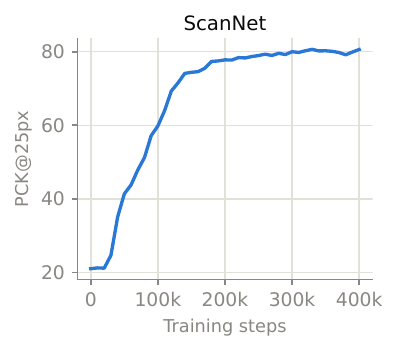}
        \caption{\textbf{Training dynamics.} Multi-view correspondence estimation attention performance.}
        \label{fig:training-dynamics}
    \end{minipage}
    \hfill
    \begin{minipage}[t]{0.64\linewidth}
        \centering
        \includegraphics[width=\linewidth]{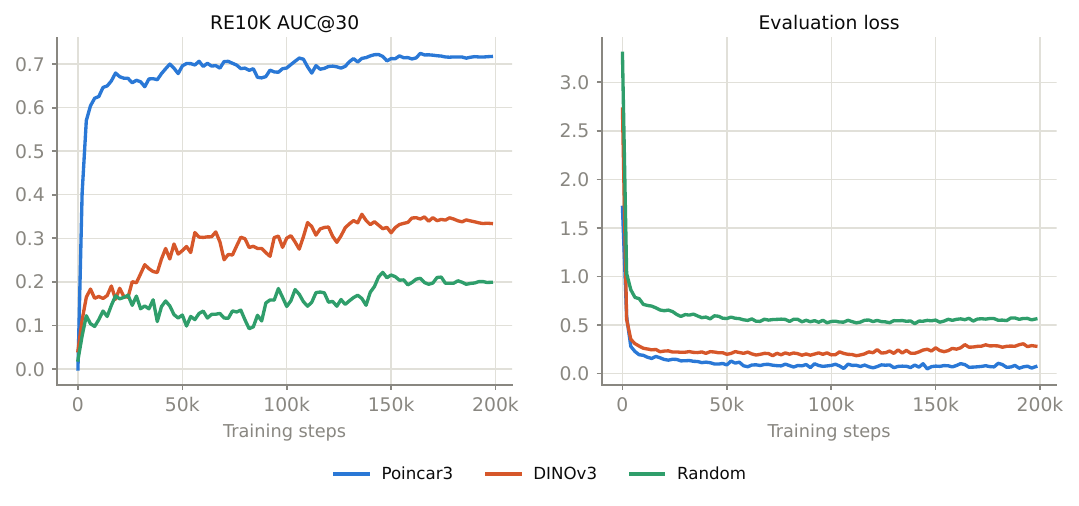}
        \caption{\textbf{Initialization for feed-forward reconstruction.} Comparing training from scratch (green), initializing the encoder from DINOv3 (red), and finetuning \ours~(blue).}
        \label{fig:training-curve-ffr}
    \end{minipage}
\end{figure}

We begin by considering feed-forward reconstruction~\citep{wang2024dust3r,wang2025vggt}, where a multi-view transformer directly predicts scene geometry from a sequence of images. We consider three evaluation settings of increasing computational complexity (i) train only the pose and depth head, (ii) train a transformer, and (iii) finetune the full model with new heads. We train on a large collection of labeled 3D datasets using 4 H200 GPUs for each experiment and report the results in \cref{tab:ffrecon} together with training curves in \cref{fig:training-curve-ffr}. \ours~shows substantially improved performance, especially on camera pose estimation, compared to all existing models. Furthermore, initializing from the \ours~weights allows for rapid finetuning, achieving an AUC@30$^\circ$ of 65.0\%+ on RE10K after only 10K steps with a low learning rate. Further details can be found in \cref{appendix:evaluation} in the Appendix.

\begin{table*}[t]
\centering
     \caption{
\textbf{Feed-forward reconstruction.} Reporting relative pose accuracy by AUC over 10 random frames and point cloud accuracy by median accuracy (acc.) in mm and normal consistency (NC) by the cosine of the angle between the normals. Training with 4 H200 GPUs for 3 days each.
\label{tab:ffrecon}} \centering
\small
\begin{tabular}{l rr rr rr rr rr}
\toprule

& \multicolumn{6}{c}{Multi-View Relative Pose} & \multicolumn{4}{c}{Point Cloud Estimation} \\

\cmidrule(lr){2-7}
\cmidrule(lr){8-11}
Method &

\multicolumn{2}{c}{RE10K} & \multicolumn{2}{c}{ScanNet++} & \multicolumn{2}{c}{MegaDepth}  & 
\multicolumn{2}{c}{ETH3D} & \multicolumn{2}{c}{DTU} \\
\cmidrule(lr){2-3}
\cmidrule(lr){4-5}
\cmidrule(lr){6-7}
\cmidrule(lr){8-9}
\cmidrule(lr){10-11}
Metric~$\rightarrow$ 
& @3$^\circ$
& @30$^\circ$

& @3$^\circ$
& @30$^\circ$

& @3$^\circ$
& @30$^\circ$

& Acc.
& NC 

& Acc. 
& NC

\\

\midrule

\rowcolor{gray!25}  \multicolumn{11}{l}{\textit{\textbf{Train only heads on top of the frozen backbone}}} \\

DINOv3 & 0.0 & 18.9 & 0.0 & 9.4 & 1.5 & 55.5 & 1.18 & 0.59 & 10.76 & 0.54 \\
Muskie & 1.1 & 36.3 & 0.0 & 19.0 & 0.1 & 46.9 & 1.12 & 0.60 & 11.43 & 0.55 \\
MuM & 0.0 & 29.8 & 0.0 & 18.7 & 0.1 & 48.5 & 1.12 & 0.60 & 12.60 & 0.54 \\
\ours & \bfseries 2.4 & \bfseries 51.9 & \bfseries 0.1 & \bfseries 48.7 & \bfseries 3.7 & \bfseries 67.5 & \bfseries 0.82 & \bfseries 0.65 & \bfseries 10.60 & \bfseries 0.59 \\

\rowcolor{gray!25}  \multicolumn{11}{l}{\textit{\textbf{Train a transformer on top of the frozen backbone}}} \\
DINOv3 & 1.0 & 29.1 & 0.0 & 22.0 & 1.7 & 65.6 & 0.95 & 0.66 & 10.63 & 0.58 \\
Muskie & 1.8 & 41.5 & 0.0 & 28.1 & 1.7 & 58.9 & 0.88 & 0.67 & 11.95 & 0.58 \\
MuM & 1.4 & 40.1 & 0.0 & 30.2 & 1.9 & 60.8 & 1.00 & 0.60 & 12.26 & 0.56 \\
\ours & \bfseries 5.2 & \bfseries 62.2 & \bfseries 1.3 & \bfseries 52.1 & \bfseries 5.2 & \bfseries 69.5 & \bfseries 0.75 & \bfseries 0.68 & \bfseries 10.52 & \bfseries 0.61 \\

\rowcolor{gray!25}  \multicolumn{11}{l}{\textit{\textbf{Finetune the backbone}$^1$}} \\

Random init. & 0.0 & 17.4 & 0.0 & 11.3 & 0.0 & 35.9 & 1.24 & 0.60 & 12.34 & 0.56\\
DINOv3 init. & 1.3 & 31.2 & 0.0 & 30.9 & 6.9 & 73.8 & 0.98 & 0.68 & 10.87 & 0.59 \\
\ours & \bfseries 8.3 & \bfseries 68.2 & \bfseries 4.5 & \bfseries 67.0 & \bfseries 7.4 & \bfseries 74.6 & \bfseries 0.81 & \bfseries 0.80 & \bfseries 6.73 & \bfseries 0.63 \\

\bottomrule
\multicolumn{11}{l}{\scriptsize $^1$MuM and Muskie are excluded because their architectures differ substantially, making direct finetuning comparisons less meaningful.} \\
\end{tabular}
\end{table*}

\subsection{Multi-view Correspondence Estimation} \label{subsec:mv-match}

\begin{figure}[t]
\centering
\includegraphics[width=0.91\linewidth]{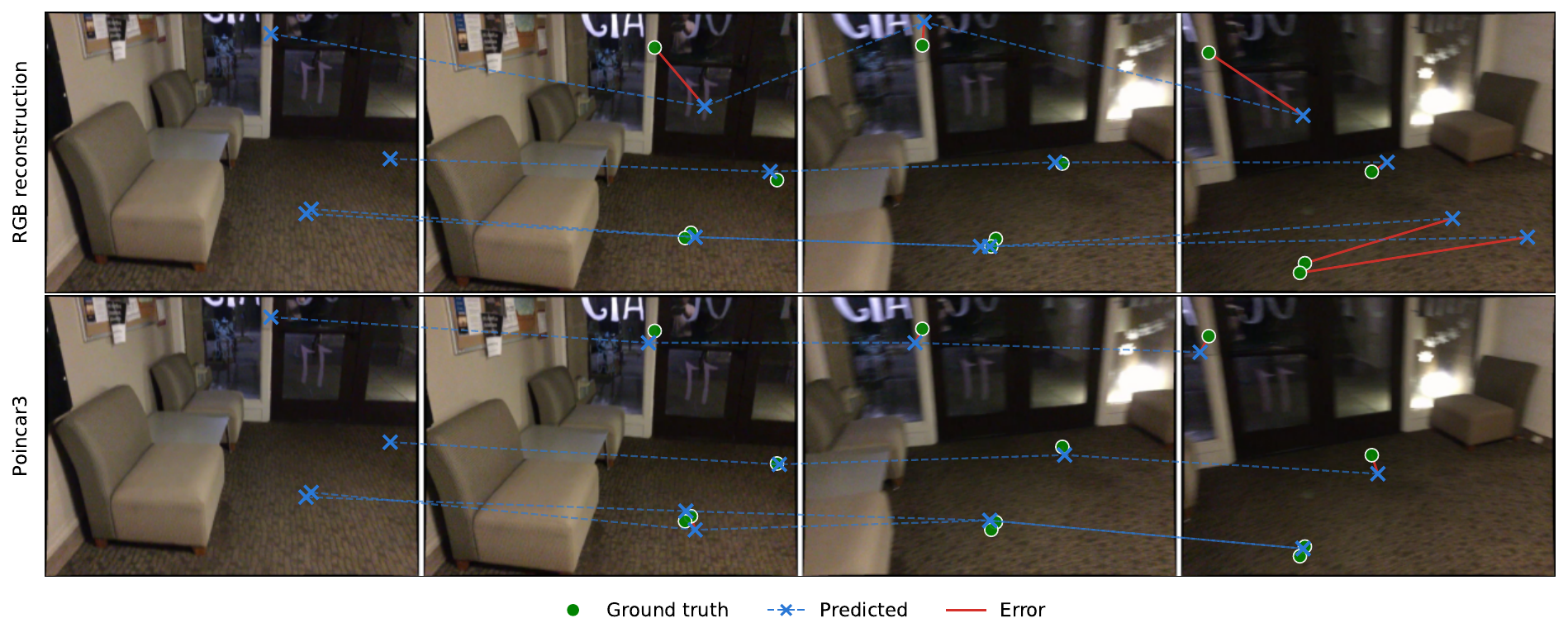}
\caption{\textbf{Multi-view correspondence tracks.} We visualize the predicted tracks (blue), the ground-truth (green), and the error (red). \ours~gives accurate zero-shot multi-view geometry.}
\label{fig:mv-match}
\end{figure}

We study the 3D geometric consistency of features by zero-shot multi-view correspondence estimation. Correspondence estimation is at the heart of multiple-view geometry~\citep{Hartley_Zisserman_2004} and has been shown to be closely linked to understanding 3D~\citep{banani2024probing3dawarenessvisual,stary2025understanding,chen2026seese3emergence3dspace}. Concretely, we follow the evaluation protocol of \citet{banani2024probing3dawarenessvisual} and sample sequences of 8 images where a set of patches are queried and tracks are produced by either nearest-neighbor matching in feature space or by using the maximum patch activation of the attention map. As shown in \cref{fig:mvcorr-layerwise} (Appendix), the performance of baselines can vary substantially per layer, although \ours~consistently outperforms at each depth. For fair comparison, we sweep the layers and report the best performing one in \cref{tab:mvcorr}. Aside from the quantitative evaluation, in \cref{fig:mv-match} we qualitatively compare the predicted tracks with the ground-truth. We find that \ours~is substantially more accurate than other SSL methods and feed-forward reconstruction models trained with 3D annotations. The attention map is an even more powerful correspondence estimator, achieving a top-accuracy of 94.9. We also outperform all other methods on two-view correspondence estimation using a linear probe (\cref{tab:matchbench}, Appendix).

\begin{table}[t]
  \centering
       \caption{
  \textbf{Multi-view correspondence estimation.} Zero-shot patch tracking across 8
  views. While \ours~has strong representations at all layers, we compare to the
  \textbf{best} layer for fair comparison.}
  \label{tab:mvcorr}
  \small
  \begin{tabular}{l rrrr rrrr}
  \toprule
  Method &
  \multicolumn{4}{c}{ScanNet~\citep{dai2017scannet}} &
  \multicolumn{4}{c}{NAVI~\citep{jampani2023navi}}  \\
  \cmidrule(lr){2-5}
  \cmidrule(lr){6-9}
  PCK@~$\rightarrow$
  & 5px & 10px & 25px & 50px
  & 5px & 10px & 25px & 50px
  \\

  \midrule

  \multicolumn{9}{l}{\textbf{Feature Nearest-Neighbor Matching}}\\
  \cmidrule(lr){1-9}

  \rowcolor{gray!25}  \multicolumn{9}{l}{\textit{\textbf{Feed-Forward
  Reconstruction}}} \\
  VGGT-$\Omega$ & 24.5 & 32.8 & 52.9 & 69.0 & 16.8 & 25.9 & 51.6 & 70.9 \\
  DA3 & 21.5 & 22.9 & 28.2 & 35.2 & 14.6 & 18.6 & 30.5 & 46.9 \\
  $\pi^3$ & 26.7 & 38.6 & 61.3 & 75.6 & 18.2 & 30.3 & 60.2 & 78.8 \\

  \rowcolor{gray!25}  \multicolumn{9}{l}{\textit{\textbf{Self-Supervised Models}}}
  \\
  DINOv3 & 24.7 & 33.9 & 56.8 & 72.2 & 17.6 & 28.7 & 59.6 & 78.7 \\

  Muskie & 27.7 & 42.4 & 70.1 & 83.1 & 18.7 & 32.7 & 64.7 & 80.5 \\
  MuM & 27.0 & 40.5 & 66.9 & 79.9 & 17.9 & 30.2 & 60.2 & 74.9 \\
  \ours & \bfseries 27.9 & \bfseries 46.2 & \bfseries 79.5 & \bfseries 89.6 &
  \bfseries 19.2 & \bfseries 33.7 & \bfseries 68.7 & \bfseries 84.2 \\

  \midrule
  \multicolumn{9}{l}{\textbf{Attention Matching}}\\
  \cmidrule(lr){1-9}

  \rowcolor{gray!25}  \multicolumn{9}{l}{\textit{\textbf{Feed-Forward
  Reconstruction}}} \\
  VGGT-$\Omega$ & 24.3 & 32.9 & 54.2 & 68.9 & 16.3 & 23.9 & 46.8 & 63.7 \\
  DA3 & 21.4 & 23.2 & 28.2 & 34.0 & 14.8 & 19.1 & 31.6 & 46.7 \\
  $\pi^3$ & \bfseries 27.0 & 40.7 & 68.7 & 83.8 & 18.3 & 30.4 & 62.2 & 82.0\\

  \rowcolor{gray!25}  \multicolumn{9}{l}{\textit{\textbf{Self-Supervised Models}}}
  \\
  Muskie & 24.6 & 34.7 & 62.9 & 81.5 & 15.5 & 22.3 & 46.2 & 67.9  \\
  MuM & 25.6 & 37.2 & 64.7 & 77.2 & 17.4 & 28.5 & 56.6 & 68.7 \\
  \ours & \bfseries 27.0 & \bfseries 43.9 & \bfseries 83.7 & \bfseries 94.9 &
  \bfseries 19.3 & \bfseries 34.7 & \bfseries 74.5 & \bfseries 86.5 \\

  \bottomrule
  \end{tabular}
  \end{table}

\subsection{Emergent SE(3) Structure}\label{subsec:poincare}

Next, we follow \citet{chen2026seese3emergence3dspace}\footnote{Reimplemented based on the paper. No code is publicly available as of writing.} and evaluate whether the features capture the \textit{geometry} of $\mathrm{SE(3)}$. Consider a static scene observed at poses $P_t\in \mathrm{SE(3)}$, and let $\mathcal{H}_t=f_\theta(\mathcal{I}_t)\in\mathbb{R}^C$ denote a model's feature for frame $\mathcal{I}_t$. The motion between frames $t$ and $t{+}s$ is the
$\mathfrak{se}(3)$ twist
\begin{equation}
\Delta P_{t,s}=\big(\mathrm{Log}(P_t^{-1}P_{t+s})\big)^{\vee}\in\mathbb{R}^6,
\end{equation}
stacking three rotational and three translational velocities. We study if $\Delta P$ is a function of the corresponding feature displacement by fitting a
trainable adapter $\varphi_\phi$ (an MLP) in a Siamese manner:
\begin{equation}
\Delta P_{t,s}\;\approx\;W\big(\varphi_\phi(\mathcal{H}_{t+s})-\varphi_\phi(\mathcal{H}_t)\big)
\label{eq:poincare}
\end{equation}
The network $\varphi_\phi$ is called a \textit{Poincar\'e adapter} as it aims to unroll the non-linear feature space to a homogeneous coordinate system, making changes in pose linear, revealing the $\mathrm{SE(3)}$ geometry.

\begin{figure}[t]
    \centering
    \hfill
    \begin{minipage}[t]{0.54\linewidth}
        \centering
        \vspace{0pt}
        \captionof{table}{\textbf{Poincar\'e adapter.} We evaluate the frozen features using a lightweight camera motion adapter on 20 held-out scenes
        from ScanNet++. Inclusion of multi-view information is denoted by \cmark.}
        \small
        \setlength{\tabcolsep}{3.5pt}
        \begin{tabular}{llrrr}
            \toprule
            Method & MV & Avg. $R^2$ & $R^2>0$ & $R^2>0.3$ \\
            \midrule
            DINOv3 & \xmark
                & 0.046\std{0.003}
                & 42.5\std{1.9}
                & 0.9\std{0.6} \\
            \midrule
            Muskie & \xmark
                & 0.042\std{0.003}
                & 39.7\std{2.2}
                & 0.4\std{0.2} \\
            Muskie & \cmark
                & 0.061\std{0.003}
                & 46.3\std{1.7}
                & 3.6\std{1.0} \\
            \midrule
            MuM & \xmark
                & 0.059\std{0.002}
                & 43.1\std{1.8}
                & 4.5\std{0.6} \\
            MuM & \cmark
                & 0.081\std{0.002}
                & 51.7\std{1.8}
                & 6.5\std{0.8} \\
            \midrule
            \ours & \xmark
                & 0.053\std{0.003}
                & 42.4\std{2.3}
                & 2.0\std{0.7} \\
            \ours & \cmark
                & \textbf{0.098}\std{0.002}
                & \textbf{55.7}\std{1.7}
                & \textbf{7.1}\std{0.7} \\
            \bottomrule
        \end{tabular}
        \label{tab:poincare}
    \end{minipage}
    \hfill
    \begin{minipage}[t]{0.44\linewidth}
        \centering
        \vspace{0pt}
        \includegraphics[width=\linewidth]{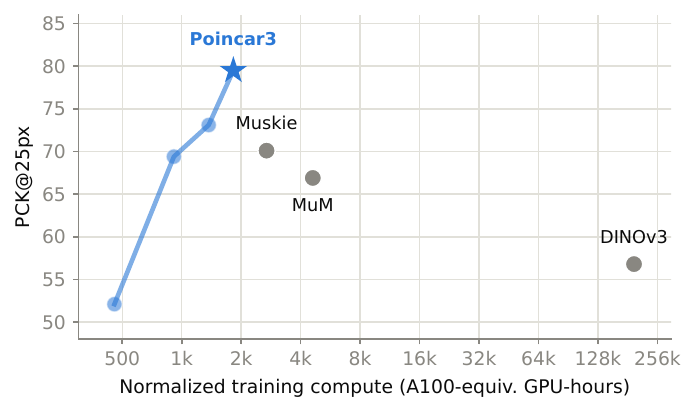}
        \caption{\textbf{Multi-view correspondence estimation accuracy versus training compute.} \ours~ties DINOv3's performance after just one day on $8\times$H200s.}
        \label{fig:pareto}
    \end{minipage}

\end{figure}

We fit one such adapter per scene on $20$ held-out ScanNet++~\citep{yeshwanth2023scannet++} scenes and report $R^2$ on the held-out pairs from each scene in \cref{tab:poincare}. We qualitatively illustrate the results in \cref{fig:poincare}. In the spirit of Poincar\'e's observation that spatial structure
emerges through distinguishing changes of position from changes of state~\citep{poincare1905value}, the motion-informed features are consistently better than those obtained from one view,
with \ours~achieving the strongest results.

\subsection{Ablation Studies}\label{subsec:ablations}

\begin{figure}[t]
    \centering

    \hfill
    \begin{minipage}[t]{0.59\linewidth}
        \centering
        \vspace{0pt}
        \captionof{table}{\textbf{Supervision ablation.}
        Multi-view correspondence estimation accuracy (PCK@25).}
        \label{tab:ablation}

        \small
        \begin{tabular}{lrr}
            \toprule
            Method &
            \multicolumn{2}{c}{Multi-view Matching} \\
            \cmidrule(lr){2-3}
            Dataset~$\rightarrow$ & ScanNet & NAVI \\
            \midrule
              \multicolumn{3}{l}{\textit{Same compute, data, and architecture}} \\
            \rowcolor{yellow!25}
            \RN{1}: RGB reconstruction (MuM obj.)\!\!\!\! & 54.5 & 46.3 \\
            \rowcolor{gray!25}
            \RN{2}: Baseline (DINOv2 obj.) & 47.2 & 35.7 \\
            \RN{3}: Multi-view DINOv2 & 49.7 & 35.9 \\
            \RN{4}: Multi-view iBOT & 55.7 & 46.4 \\
            \RN{5}: +Image-level objective & 66.7 & 58.1 \\
            \RN{6}: +Teacher seeing more views & 70.2 & 61.5 \\
            \midrule
            \rowcolor{green!25}
            \RN{7}: +Increase scale (\ours)
            & \bfseries 83.7 & \bfseries 74.5 \\
            \bottomrule
        \end{tabular}
    \end{minipage}
    \begin{minipage}[t]{0.37\linewidth}
        \centering
        \vspace{0pt}
        \includegraphics[width=\linewidth]{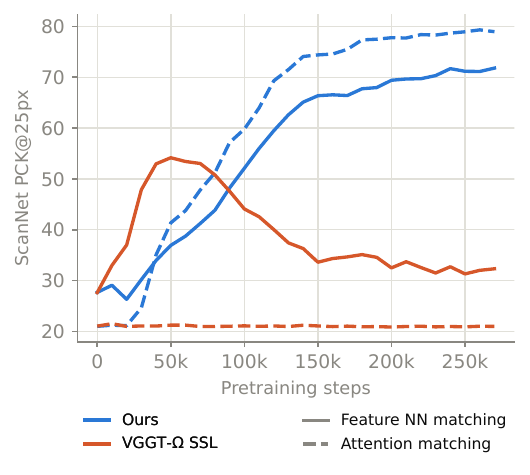}
        \caption{\textbf{VGGT-$\Omega$ SSL} comp.}
\label{fig:ablation-vggt-omega-ssl}
    \end{minipage}

\end{figure}

\paragraph{Supervision.} In \cref{tab:ablation}, we compare different supervision objectives. For fair comparison, we keep the architecture fixed throughout and fix the compute budget to $4\times$H200 for a day. We establish two baselines: \RN{1}) multi-view RGB reconstruction and \RN{2}) single-view DINOv2. Building on the DINOv2 objective, we extend it to multiple views (\RN{3}), remove the local crops (\RN{4}), add a global objective (\RN{5}), feed the teacher additional views (\RN{6}), and scale the compute and model size (\RN{7}). Our objective (\RN{5}) is clearly superior to the RGB reconstruction used in MuM and Muskie (\RN{1}). Finally, in \cref{fig:ablation-vggt-omega-ssl} we find the MSE loss from VGGT-$\Omega$ (SSL) does not work from scratch as self-supervision.

\paragraph{Data.} Our setup enables us to leverage internet-scale data inaccessible to 3D reconstruction models requiring 3D labels. To quantify its benefit, we compare multi-view correspondence accuracy from attention maps when training on only 3D-labeled data versus all available data. Incorporating the additional data substantially improves multi-view geometry understanding, increasing PCK@50 from 87.5 to 94.9 on ScanNet and from 78.5 to 86.5 on NAVI.

\section{Limitations}

Our approach prioritizes geometric over semantic performance (see \cref{tab:semantic}, Appendix). This is consistent with evidence that semantic and spatial information are processed by different neural systems in the human brain~\citep{Binder2009Where,Epstein2017CognitiveMap}. Furthermore, we do not have the computational resources to match the size of DINOv3 (we use around 100x less compute). Yet, we significantly outperform previous approaches (see \cref{fig:pareto}). 

We find that the representations degrade when hyper-parameters such as the EMA coefficient and weight decay deviate from those of DINOv3. We also find it crucial to follow prior work when adapting the learning rate to batch size~\citep{goyal2018accuratelargeminibatchsgd}. This fragility has plagued self-distillation since its inception~\citep{grill2020bootstrap,caron2021dino} and mitigating it constitutes interesting future work. Finally, while our training is self-supervised, frame selection within video sequences is currently handcrafted. Automating this selection is an interesting direction for future work.

\section{Conclusions}

We introduced \ours, a multi-view foundation model trained from scratch in a self-supervised manner. To remove the RGB reconstruction supervision used in MuM and Muskie, which entangles geometry with appearance, we proposed a DINO-like self-distillation objective. Self-distillation is notoriously challenging to train, and introducing a multi-view transformer further exacerbates this difficulty. We show that a multi-view teacher--student objective can nevertheless be trained from scratch, without supervision, by feeding the teacher additional frames, using a global stabilizing objective, and retaining full images rather than relying on local and global crops. \ours~improves substantially over MuM and Muskie on 3D vision tasks and achieves higher performance than DINOv3 using $100\times$ less compute. These results show that multi-view self-distillation can learn strong 3D representations from scratch without explicit 3D supervision or pixel reconstruction.

\section*{Acknowledgements}
This work was supported by the Wallenberg Artificial
Intelligence, Autonomous Systems and Software Program
(WASP), funded by the Knut and Alice Wallenberg Foundation, and by the strategic research environment
ELLIIT, funded by the Swedish government.  The computational resources were provided by the
National Academic Infrastructure for Supercomputing in
Sweden (NAISS) at C3SE, partially funded by the Swedish Research Council through grant agreement no.~2022-06725, and by the Berzelius resource, provided by the Knut and Alice Wallenberg Foundation at the National Supercomputer Centre.

This work benefited from Hi! PARIS and State funding managed by the French National Research Agency (ANR) under the France 2030 program, reference ANR-23-IACL-0005. 

This work was supported by the Bosch Research Foundation (Bosch Forschungsstiftung) and by the European Union (ERC Advanced Grant Explorer, Funding ID \#101097259). %

\bibliography{iclr2027_conference}
\bibliographystyle{iclr2027_conference}

\clearpage
\appendix

\section{Things That Did Not Work} \label{sec:supp:not-work}

While not included in the ablation in the main paper (\cref{tab:ablation}), we experimented with multiple SSL objectives before arriving at our formulation. We primarily sought a simpler objective, such as BYOL~\citep{grill2020bootstrap}, which does not use centering or Sinkhorn-Knopp. While showing initial promise, the feature maps quickly degraded during training, resulting in unsatisfactory qualitative results. We attempted to mitigate this degradation through regularization, but were unable to obtain stable training. Neither centering nor Sinkhorn--Knopp normalization, nor more sophisticated regularization methods such as VICReg~\citep{bardes2022vicreg}, resolved the issue.

We also tried being more sloppy with our data curation. However, in contrast to RGB reconstruction, self-distillation appears to be more sensitive to low-quality data samples. This is likely why the DINO-team has spent significant effort in creating elaborate data curation pipelines~\citep{vo2024automaticdatacurationselfsupervised}.

\section{Baselines}\label{sec:appendix:baselines}

In the paper, our core comparisons are to the features obtained by state-of-the-art SSL methods. Namely, DINOv3, MuM, and Muskie. We omit the comparison to CroCo, which was already shown to be inferior to MuM and Muskie in their respective papers. Furthermore, MuM showed a drastic improvement over VJEPA 2 on geometric understanding. At times, we compare to the features obtained by supervised 3D models such as VGGT-$\Omega$, $\pi^3$, and DepthAnything3 (DA3). These constitute the state-of-the-art supervised methods and have garnered widespread interest from the 3D vision community. We include these as comparisons where they constitute a fair comparison, as in correspondence estimation, whereas in others, such as feed-forward reconstruction, their training objective is tailored for that specific task and are thus omitted. Lastly, we do not compare to GenCeption, as it is diffusion-based, or Alligat0r, as it requires covisibility labels.

\section{Additional Experiments}\label{sec:appendix:experiments}

\subsection{Qualitative }

\begin{figure}[t]
\centering
\includegraphics[width=0.99\linewidth]{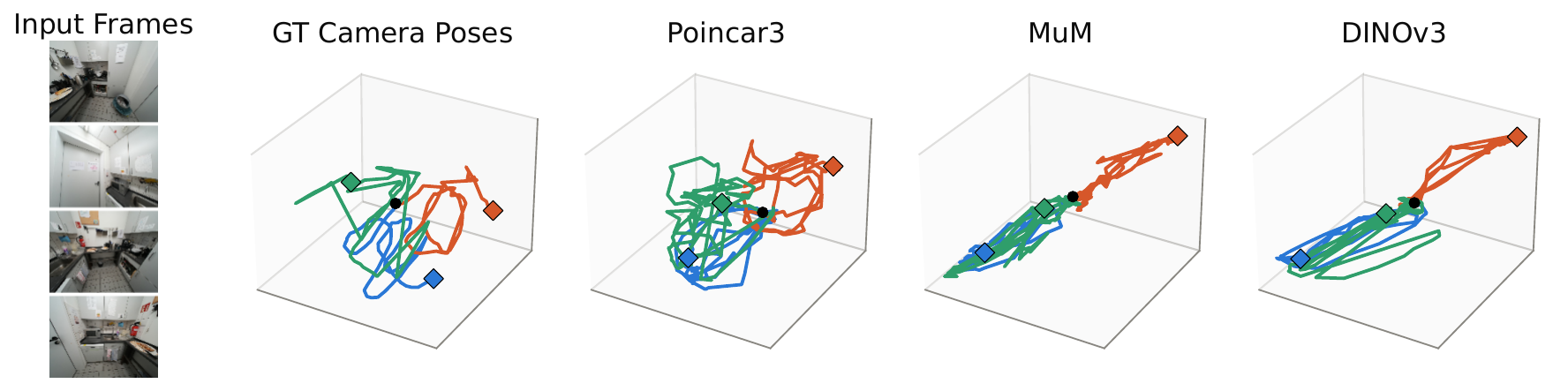}
\caption{\textbf{Poincar\'e adapter.} We visualize the alignment between visual features and camera trajectories with a lightweight \textit{Poincar\'e adapter}, following Fig.~1 of \citet{chen2026seese3emergence3dspace}. Applied to \ours, it unrolls the feature space toward the ground-truth camera trajectories.}
\label{fig:poincare}
\end{figure}

\begin{figure}[t]
\centering
\includegraphics[width=0.99\linewidth]{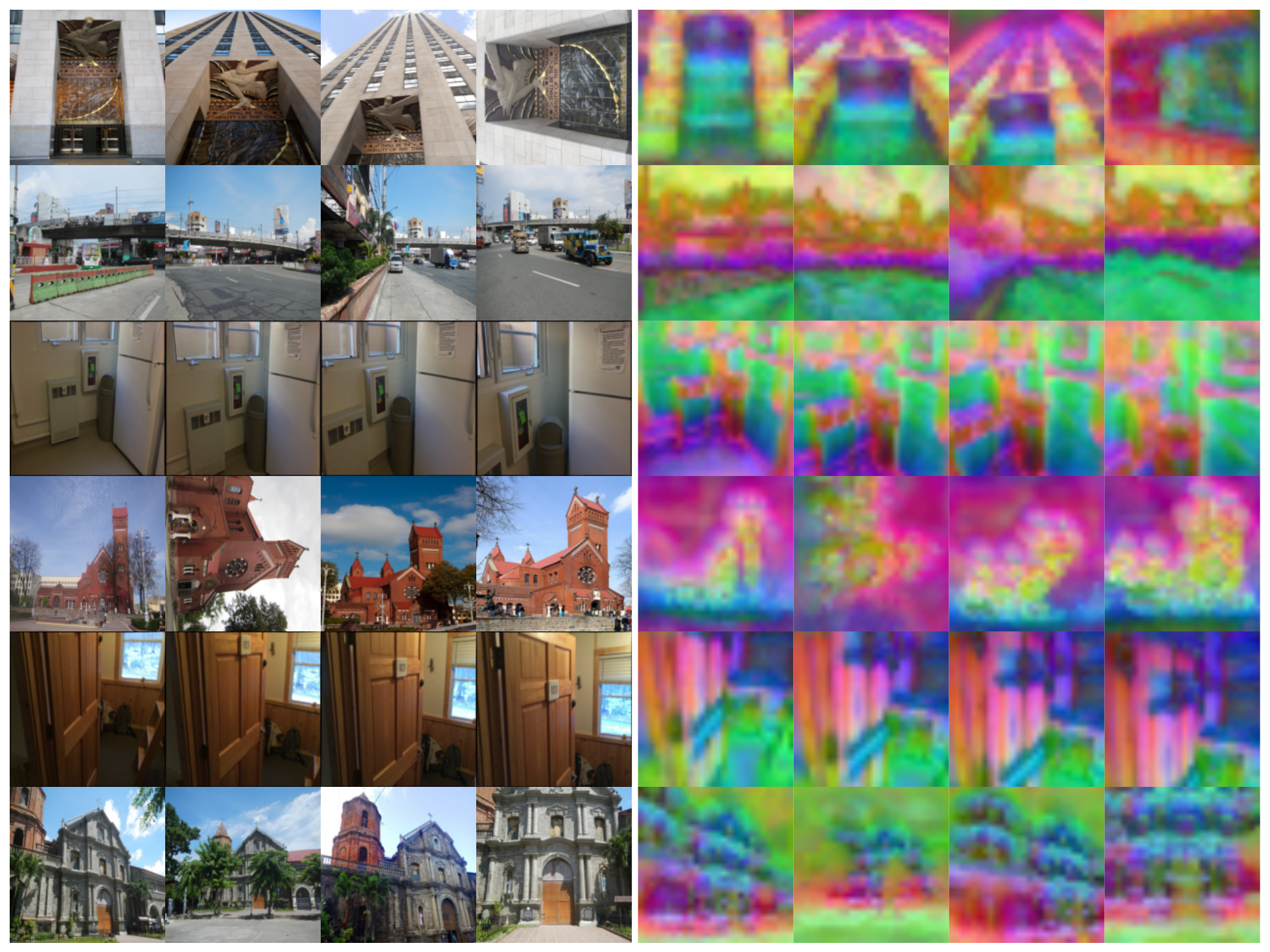}
\caption{\textbf{PCA feature visualization.} We visualize the 3 principal components as RGB for different sequences from the same scene.}
\label{fig:pca}
\end{figure}

\begin{figure}[t]
\centering
\includegraphics[width=0.99\linewidth]{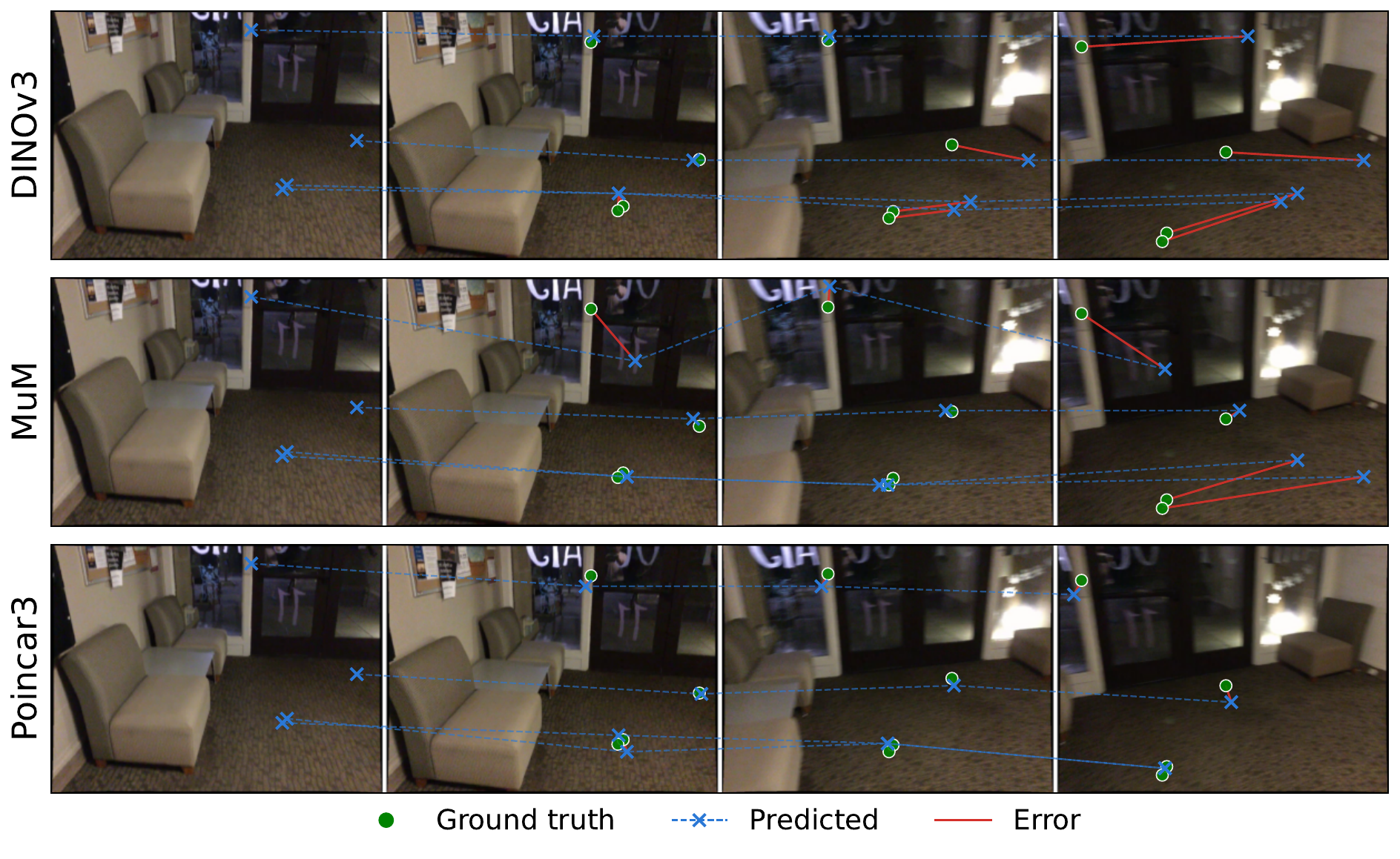}
\caption{\textbf{Additional multi-view correspondence visualizations.} Including also DINOv3 as a baseline.}
\label{fig:mv-corr-more}
\end{figure}

To understand what the model has learned, we visualize the PCA of the patch features after the decoder. We show the results for an image pair in \cref{fig:pca}. Furthermore, we visualize the alignment of the Poincar\'e adapter with the ground-truth camera trajectory in \cref{fig:poincare} and additional baselines for qualitative multi-view correspondence tracks in \cref{fig:mv-corr-more}.

\subsection{Two-view Matching} \label{subsec:matching}

\begin{table}[t]
\centering
     \caption{
\textbf{Lightweight two-view matching.} Matching robustness on ScanNet-1500~\citep{dai2017scannet,sarlin2020superglue}.}\label{tab:matchbench} \centering
\small
\begin{tabular}{l rrr rrr}
\toprule
Method &

\multicolumn{3}{c}{Nearest-Neighbor} & \multicolumn{3}{c}{Linear Probe} \\
\cmidrule(lr){2-4}
\cmidrule(lr){5-7}
PCK~$\rightarrow$ 
& 8px
& 16px
& 32px

& 8px
& 16px
& 32px

\\

\midrule

\rowcolor{gray!25}  \multicolumn{7}{l}{\textit{\textbf{Feed-forward Reconstruction}}} \\
VGGT-$\Omega$ & 2.3 & 6.8 & 17.0 & 26.4 & 47.9 & 67.5 \\
DA3 & 0.8 & 2.4 & 6.8 & 29.2 & 49.0 & 66.7 \\
$\pi^3$ & 8.0 & 17.1 & 30.2 & 32.9 & 55.3 & 73.5 \\

\rowcolor{gray!25}  \multicolumn{7}{l}{\textit{\textbf{Self-supervised Models}}} \\
DINOv3 & 17.5 & 35.0 & 52.2 & 35.3 & 60.4 & 77.9 \\
Muskie & 6.6 & 16.2 & 32.1 & 34.8 & 54.5 & 69.8 \\
MuM & 15.7 & 31.9 & 51.1 & 40.7 & 63.6 & 78.9 \\
\ours & \bfseries 24.3 & \bfseries 42.5 & \bfseries 58.6 & \bfseries 45.3 & \bfseries 67.0 & \bfseries 79.3 \\

\bottomrule
\end{tabular}
\end{table}

We now consider two-view matching. We employ a lightweight protocol by training linear probes and using zero-shot nearest-neighbor (NN) in feature space. This evaluation protocol follows from \citet{edstedt2024roma} and \citet{nordstrom2026mum}. We report the accuracy in \cref{tab:matchbench} on ScanNet-1500. We find that \ours~outperforms all other methods on both zero-shot matching and linear probing. 

\subsection{Semantic Tasks}

We compare the semantic performance in \cref{tab:semantic}. Without being trained on ImageNet, \ours~outperforms other multi-view SSL models on image classification and semantic segmentation. However, removing the local crops and training on significantly more geometric data leads to significant performance degradation compared to DINOv3. For a fair comparison, we use the output of the final layer for all the models whereas in MuM, the semantic performance is evaluated after only the encoder. We follow the evaluation protocol of \citet{CAPI:2025}.

\begin{table}[t]
\centering
\caption{\textbf{Semantic performance.} Image classification accuracy on ImageNet-1K and semantic segmentation (mIoU) on ADE20K.}
\label{tab:semantic}
\small
\begin{tabular}{l rr}
\toprule
Context
& ImageNet-1K~\citep{imageNet2009,krizhevsky2012imagenet} & ADE20K~\citep{ade20k:2017} \\
\midrule

\textcolor{gray}{DINOv3} & \textcolor{gray}{85.1} & \textcolor{gray}{49.8} \\
\midrule
Muskie & 27.5 & 6.4 \\
MuM & 27.5 & 4.6 \\
\ours & \bfseries 34.4 & \bfseries 9.6 \\

\bottomrule
\end{tabular}
\end{table}

\subsection{Layer-wise Multi-View Correspondence Estimation}

\begin{figure}[t]
\centering
\includegraphics[width=0.99\linewidth]{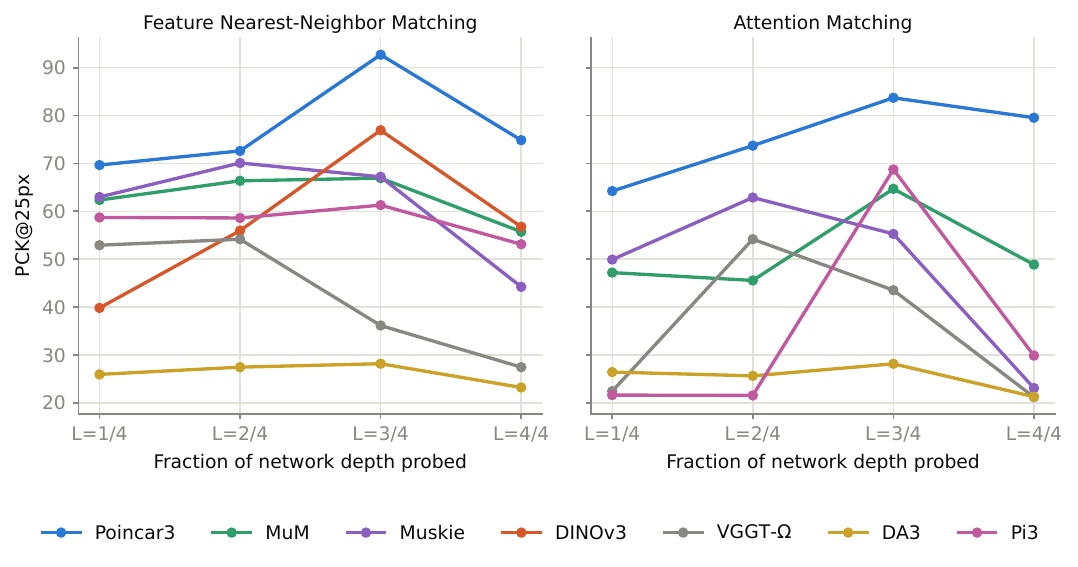}
\caption{\textbf{Layer-wise performance.} We plot the multi-view correspondence accuracy at different depths. We find that \ours~encodes multi-view geometry throughout the network, while the performance of supervised methods collapses at later layers.}
\label{fig:mvcorr-layerwise}
\end{figure}

We investigate the geometric information at different depths for the models by multi-view correspondence estimation accuracy in \cref{fig:mvcorr-layerwise}. We find that \ours~consistently outperforms the baselines at all depths.

\section{Implementation Details} \label{sec:supp:implementation-details}

\subsection{Hyper-parameters}\label{subsec:supp:hparams}

\paragraph{Architecture.} \ours{} uses a ViT-L/16 backbone with
DINOv3-style RoPE position embeddings, QK-norm, and LayerScale,
initialized from scratch. The
multi-view decoder has 12 blocks, alternating full cross-frame
attention with a cheaper register-bottleneck attention at
blocks $\{2,6,9\}$ (zero-indexed) using 16 register tokens.

\paragraph{Self-distillation objective.} We use separate
DINO/iBOT heads for the patch and global tokens, each with
65{,}536 prototypes, hidden dimension 2048, and bottleneck
dimension 256. Teacher targets are normalized with
Sinkhorn--Knopp (3 iterations) rather than centering, which we
found more resistant to collapse. The student temperature is
fixed at 0.1, while the teacher temperature warms up from 0.04
to 0.07 over the first 30k steps; the teacher EMA decay ($\lambda$)
follows the same warmup schedule, from 0.994 to 0.999. The last layer of
both heads is frozen for the first 2k steps.

\paragraph{Multi-view sampling.} Each training sequence draws a
student frame count uniformly from $[2,24]$ and, crucially,
feeds the teacher $[0,12]$ additional frames the student never
sees, unmasked; the loss is computed only over the frames both
networks observe. Patches are masked independently per frame
with probability 0.5 and a mask ratio sampled from $[0.1,
0.5]$, following iBOT. The global objective compares teacher and
student tokens frame-by-frame. Batches are built dynamically to
a fixed budget of 64 frames per GPU. Images are resized to $256\times256$ with
patch size 16. As we train on 8 H200 GPUs, the maximum number of frames can go into a batch is 512.

\paragraph{Optimization.} We train with AdamW and a fixed learning rate of $2\times10^{-4}$, warmed up over 10k
steps. Weight decay is 0.04 and
gradients are clipped to norm 1.0. Training uses bf16 mixed
precision.

\subsection{Data}

We detail our dataset mixture in \cref{tab:dataset_mix}. The majority of the scenes come from videos from the internet, providing a rich source of data that can only be mined through SSL. Note there is no overlap between the training scenes and the scenes used for evaluation in \cref{sec:results}.

\begin{table}[ht] \centering
\caption{\textbf{Dataset mixture} for {\ours }. The top part contains large-scale internet video datasets with noisy or no annotations, while the bottom part contains 3D datasets with annotations. The weight is proportional to the probability of sampling from the respective dataset.}
\label{tab:dataset_mix}
\setlength{\tabcolsep}{2pt}
\begin{tabular}{lccr}
\hline
\toprule
Datasets                          & Type / Source            & Weight & \textnumero Scenes \\ 

\midrule
SpatialVID~\citep{wang2025spatialvid}  & Outdoor / Video        &1&  176,749 \\
DL3DV~\citep{ling2024dl3dv}    & Mixed / Video        &1&  10,000 \\
RealEstate10K~\citep{realestate10k}    & Indoor / Video        &1&  7,850 \\

\midrule
MegaDepth~\citep{li2018megadepth}    & Outdoor / MVS        &1&  169 \\
AerialMD~\citep{vuong2025aerialmegadepth}    & Aerial / MVS        &1&  124 \\
BlendedMVS~\citep{yao2020blendedmvs}  & Aerial / Mesh      &1&  493 \\
Hypersim~\citep{roberts2021hypersim}& Indoor / Graphics          &1&  393 \\
TartanAir v2~\citep{wang2020tartanair}& Outdoor / Graphics          &1&  46 \\
Map-Free~\citep{arnold2022map}& Object-centric / MVS &1&  397 \\
ScanNet++ v2~\citep{yeshwanth2023scannet++} & Indoor / Mesh & 1 & 856 \\
FlyingThings3D~\citep{mayer2016large} & Outdoor / Graphics & 0.5\phantom{0} & 2,239\\
ARKitScenes~\citep{dehghan2021arkitscenes} & Indoor / RGB-D & 0.1\phantom{0} & 5,047\\
UnrealStereo4k~\citep{tosi2021unrealstereo4k} & Outdoor / Graphics & 0.01 & 8\\
Virtual KITTI 2~\citep{gaidon2016virtual,cabon2020vkitti2} & Outdoor / Graphics & 0.01 & 5\\

\midrule
\textbf{Total}                        &                     & & \textbf{204,376} \\
\bottomrule
\end{tabular}
\normalsize
\end{table}

\subsection{Training Dynamics}

\begin{figure}[t]
    \centering
    \begin{minipage}[t]{0.45\linewidth}
        \centering
        \includegraphics[width=\linewidth]{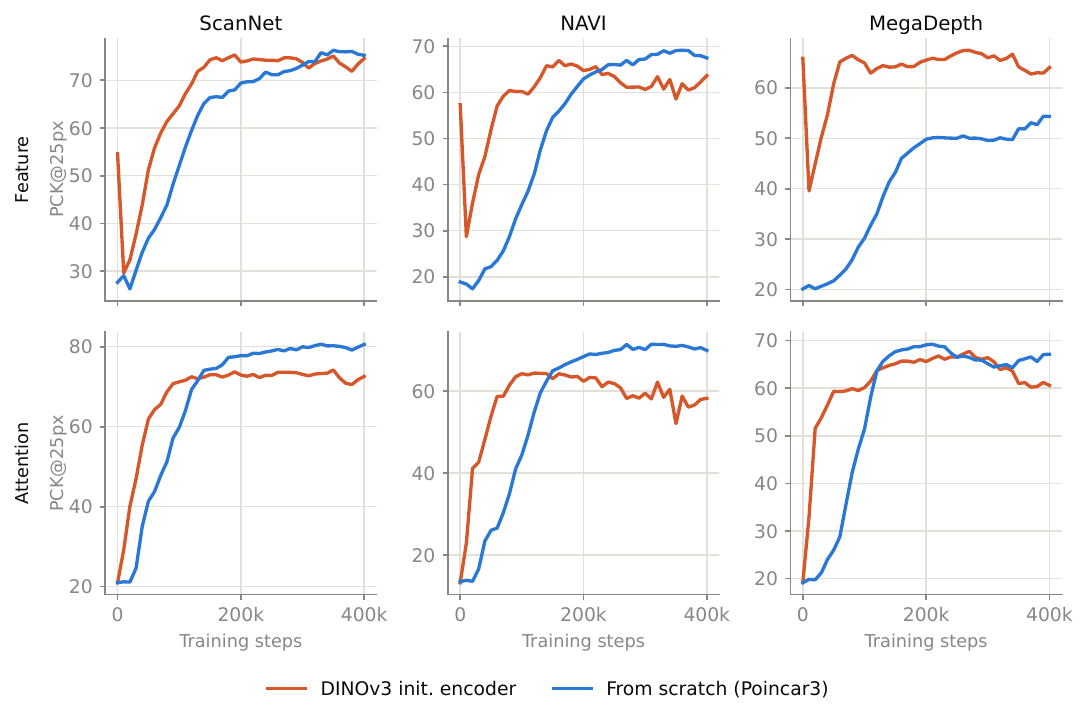}
        \caption{\textbf{Detailed evaluation metrics.} Multi-view correspondence estimation performance, from scratch vs. DINOv3 init.}
        \label{fig:training-dynamics-detailed}
    \end{minipage}
    \hfill
    \begin{minipage}[t]{0.53\linewidth}
        \centering
        \includegraphics[width=\linewidth]{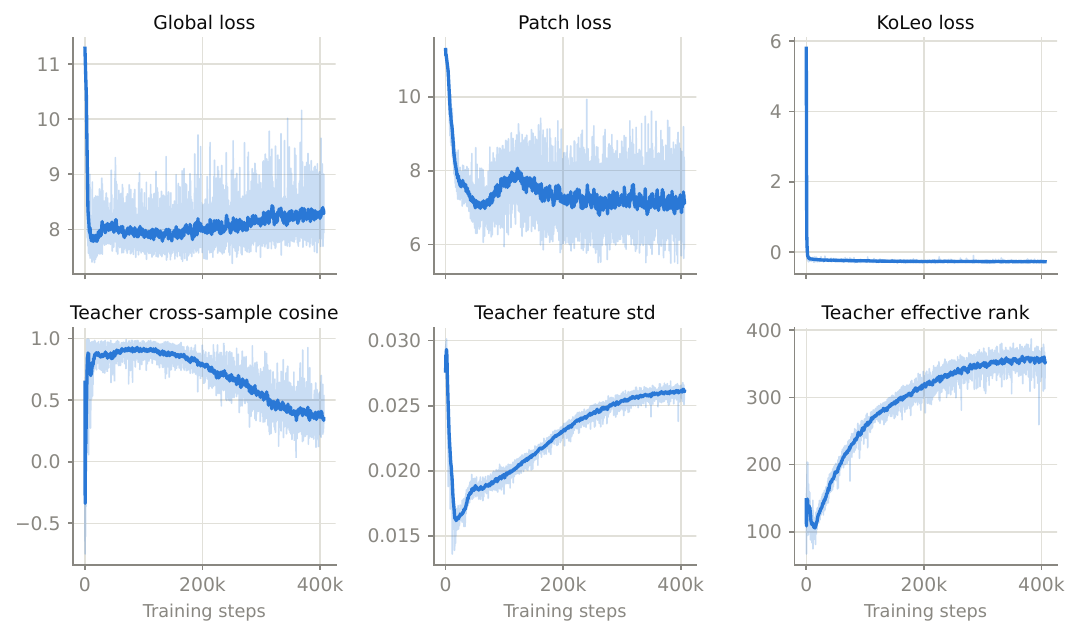}
        \caption{\textbf{Detailed stability metrics.} We visualize how the loss terms evolve during training as well as stability metrics.}
        \label{fig:training-panel}
    \end{minipage}
\end{figure}

In \cref{fig:training-dynamics-detailed}, we visualize how six different evaluation metrics evolve during training and how performance is impacted by initializing the encoder from DINOv3. Furthermore, we provide plots of the stabilization metrics we tracked throughout training in \cref{fig:training-panel}. While the losses are uninformative in a teacher--student setup, it is encouraging to see that the teacher's features are not collapsing but rather becoming more discriminative as training progresses (as illustrated by the standard deviation and the effective rank).

\section{Evaluation}\label{appendix:evaluation}

\subsection{Feed-forward reconstruction}
\label{sec:appendix-ffrecon}

We broadly follow the training in~\citet{wang2026vggtomega}, but make minor adjustments to fit with our limited compute budget. The protocol we use is outlined below.

\paragraph{Training.}
Every feed-forward reconstruction model shares the same pair of prediction heads on top of a
backbone: a \emph{camera head}, a 4-layer self-attention trunk that mixes each frame's camera and
register tokens across the whole sequence in one shot and regresses a 9D pose encoding per frame
(translation, rotation quaternion, field of view), and a \emph{depth head}, a DPT-style
multi-scale fusion decoder (following VGGT/Depth-Anything) that reads dense tokens from four
evenly-spaced depths of the backbone and regresses per-pixel depth and confidence independently
per frame (no cross-frame mixing). Training data is a weighted mixture of posed,
depth-annotated multi-view datasets (HyperSim, ScanNet++, MegaDepth, BlendedMVS, Map-free, MegaSynth, and SpatialVid). Each batch draws one sequence length
(uniform in $[2,24]$ frames) and one resolution bucket, following the same dynamic-batching
scheme used for pretraining of \ours.

We compare three training protocols that trade off how much of the backbone is updated:
\begin{enumerate}
    \item \textbf{Frozen backbone, heads only.} The backbone (encoder and, where present, its
pretrained cross-view decoder) stays entirely frozen, and only the camera/depth heads are
trained on top of its existing features, isolating how much 3D structure a backbone's
pretraining objective already encodes.

\item \textbf{Frozen backbone + adapter.} The backbone
stays frozen, but a small, freshly-initialized adapter (two layers of the same alternating
frame/inter-frame self-attention used in our own cross-view decoder) sits between the frozen
per-frame features and the heads, giving every backbone an identical amount of new trainable
cross-view capacity, which matters for backbones with no pretrained cross-view mixing at all
(e.g.\ DINOv3, a plain per-frame ViT).
\item \textbf{Full finetune.} The entire backbone is
updated jointly with the heads end to end, optionally regularized against representational drift
with a feature-distillation term against a frozen VGGT-$\Omega$ teacher, matching both dense
patch tokens and the camera token. From this experiment we do not include MuM or Muskie as it would constitute an unfair comparison as their architectures differ. For the DINOv3 comparison, we substitute the encoder for DINOv3 ViT-L/16. Muskie showed that this is superior to transplanting the DINO weights also into the multi-view transformer. 
\end{enumerate}

In all three protocols the training objective is
$\mathcal{L} = \lambda_{\text{cam}}\mathcal{L}_{\text{cam}} + \lambda_{\text{depth}}\mathcal{L}_{\text{depth}}$, with
$\lambda_{\text{cam}} = \lambda_{\text{depth}} = 1$. The camera loss is a flat L1 on the
predicted vs.\ ground-truth pose encoding $p = (t, q, f)$ (translation, rotation quaternion,
field-of-view):
\begin{equation}
    \mathcal{L}_{\text{cam}} = \| \hat p - p \|_1 .
\end{equation}
The depth loss combines a confidence-weighted term, a plain regression term, and a multi-scale
gradient term over pixels with valid ground-truth depth (after dropping the worst 2\% by
quantile):
\begin{equation}
    \mathcal{L}_{\text{depth}} = \mathbb{E}\big[\gamma\,|\hat d - d|\,c - \alpha \log c\big]
    + \mathbb{E}\big[|\hat d - d|\big] + \mathcal{L}_{\text{grad}},
\end{equation}
where $c$ is the predicted per-pixel confidence, $\gamma{=}1$, $\alpha{=}0.2$ (the
$-\alpha\log c$ term discourages trivially collapsing confidence to zero), and
$\mathcal{L}_{\text{grad}}$ averages an L1 gradient-matching term over 4 progressively
downsampled resolutions. Ground-truth depth and camera translations are rescaled per training
batch, since absolute scale is unobservable from monocular images alone.

\paragraph{Evaluation.}
We evaluate the resulting checkpoints along two axes: relative camera pose and multi-view
point-cloud geometry. For the former, we follow \citet{wang2025vggt}, while for the latter we follow \citet{wang2025pi3}. For pose, a single joint forward pass over $N{=}10$ sampled frames of a
test sequence predicts every frame's pose in one model-chosen reference frame (VGGT-style);
since this absolute frame carries no ground-truth correspondence, we score every one of the
$\binom{N}{2}$ frame pairs' \emph{relative} pose instead -- rotation error is the geodesic angle
between predicted and ground-truth quaternions, translation error is the angle between the
(sign-disambiguated) predicted and ground-truth translation directions -- and report the area
under the cumulative-accuracy curve of $\max(\text{rotation error}, \text{translation error})$,
thresholded at $30^\circ$ (AUC@30), averaged first within each test scene and then across scenes
so scenes contributing more sampled sequences (e.g.\ RE10K) do not dominate. Test sets are
MegaDepth-1500, RE10K, ScanNet-1500, and 50 held-out ScanNet++ scenes, each following its own
reference protocol's image-loading convention.

For point-cloud quality, we unproject each backbone's predicted per-frame depth through its
predicted camera parameters into a world-space point cloud, on 5-view sequences from DTU and
ETH3D. Since predicted geometry is only defined up to an unknown similarity transform, we first
coarsely align it to the ground-truth point cloud with Umeyama's method and then refine with
rigid ICP. We report point-to-point accuracy and completion (mean/median nearest-neighbor
distance predicted$\to$ground-truth and vice versa) and normal consistency (mean cosine
similarity between each point's PCA-estimated normal and its nearest neighbor's), averaged over
sequences -- reproducing VGGT's own multi-view reconstruction evaluation protocol.

\subsection{Multi-view correspondence estimation}
\label{sec:appendix-mvcorr}

Our protocol follows \citet{banani2024probing3dawarenessvisual}. In particular, we evaluate zero-shot correspondence directly, without any task-specific finetuning, by
feeding a cluster of $8$ covisible views of one scene through a backbone in a single multi-view
forward pass and tracking $100$ query points sampled in the first view across the remaining $7$. Ground-truth tracks come from reprojecting each query point
through its ground-truth depth, pose, and intrinsics into every other view, with a depth-based
occlusion test discarding views where the reprojected point is not actually visible; points never
visible in any other view are dropped entirely. We evaluate on NAVI and ScanNet (both with
metric-scale ground-truth depth) and MegaDepth (unscaled SfM depth).

We score two different ways of reading off a correspondence from the same forward pass.
\emph{Feature matching} takes the source query point's dense feature (bilinearly sampled at its
exact sub-patch location) and finds its 1-nearest-neighbor by cosine similarity in each target
view's dense feature map. \emph{Attention matching} instead reads the correspondence directly off
the backbone's own softmax cross-view attention: the source point's query vector attends over
every view's keys, and the arg-max of that distribution restricted to a target view's patch
tokens is the match -- this asks whether the model's own attention agrees with correspondence,
not just its output feature similarity. Both are read from features/attention at a specific
network depth; since backbones differ widely in depth and in which blocks carry genuine
cross-view mixing, we sweep four evenly-spaced relative depths (quarter-points of each backbone's
own valid block range) and, following common practice, report the single best-performing depth
per backbone and correspondence method.

Given a predicted match and its ground-truth target location, we report the fraction of visible
query points landing within $k$ pixels of the ground-truth match (PCK@$k$, $k \in \{5, 10, 25,
50\}$), plus the analogous 3D accuracy after unprojecting both points with ground-truth depth.

\subsection{Emergent SE(3) Structure} 

We follow \citet{chen2026seese3emergence3dspace}. As the code is unpublished as of writing, we do our best to reimplement the core protocol. In particular, we fit a Poincar\'e adapter per scene on $20$ held-out ScanNet++~\citep{yeshwanth2023scannet++} scenes, using a chronological $80/20$ split of each scene's frames and sweeping strides $s\in\{2,\dots,60\}$, and measure $R^2$ on the held-out pairs. As $R^2$ is unbounded below, its raw mean is dominated by a few degenerate scenes; we therefore report the clipped mean $\overline{R^2}=\mathrm{mean}\,(\max(0,R^2))$ together with the fraction of scenes exceeding $R^2\!>\!0$ and $R^2\!>\!0.3$. All values are the best over four relative encoder depths and are averaged over $10$ adapter seeds ($\pm$ 95\%-confidence interval where the scene set is fixed and shared across methods). Methods marked MV additionally receive an eight-frame context window with a frame spacing of $5$, activating their cross-view attention; only the anchor frame's feature is retained.

\subsection{Visualizing feature correlations}

Here we detail how we perform a visualization such as \cref{fig:feature-corr}. Given a cluster of covisible frames processed jointly in one multi-view forward pass, we mark a query patch in the first frame and, for every other frame, compute the cosine similarity between the query's dense feature and every patch of that frame (clipped below zero). Each row (backbone) is independently rescaled to its own $5^{\text{th}}$-percentile--to-max range and rendered as a semi-transparent Turbo-colormap overlay (opacity $0.25$--$0.85$, so weakly-correlated regions stay visible rather than fading to the bare image), with a circle marking the arg-max match. Rescaling per backbone sacrifices strict cross-row comparability of absolute intensity in exchange for making each backbone's correlation \emph{dispersion} legible: a backbone that discriminates well shows a small, concentrated hot region, while one that does not washes warm over most of the frame.

\end{document}